%% file: seq2seq.tex
\documentclass[a4paper,11pt]{article}

\usepackage{amsmath,amssymb,amsthm,bbm,bm}  
\usepackage{authblk}  
\usepackage[british]{babel}
\usepackage{balance}
\usepackage[sorting=none,doi=false,isbn=false,giveninits=true,style=nature,url=false]{biblatex}           
\usepackage{booktabs} 
\usepackage{caption}
\usepackage{color}
\usepackage{comment}
\usepackage{enumitem}
\usepackage[T1]{fontenc}
\usepackage[top=2.5cm, bottom=2.5cm, left=2.5cm, right=2.5cm]{geometry}
\usepackage{graphicx}
\PassOptionsToPackage{hyphens}{url}  
\usepackage[colorlinks = true,
            linkcolor = OliveGreen,
            urlcolor  = blue,
            citecolor = MidnightBlue,
            anchorcolor = blue]{hyperref}
\usepackage{cleveref}  
\usepackage[utf8]{inputenc}
\usepackage{csquotes}  
\usepackage[mono=false]{libertine}  
\usepackage{mathtools}
\usepackage{microtype}      
\usepackage{nicefrac}
\usepackage{setspace}
\usepackage{subfigure}
\usepackage[usenames,dvipsnames]{xcolor}

\input{settings}

\newcommand{\citet}[1]{\textcite{#1}}
\bibliography{references}

\title{The impact of memory on learning sequence-to-sequence tasks 
}

\author[1]{Alireza Seif\thanks{\texttt{seif@uchicago.edu}}}
\author[2]{Sarah A.M.~Loos}
\author[3]{Gennaro Tucci}
\author[2]{{\'E}dgar Rold{\'a}n}
\author[4]{Sebastian Goldt\thanks{\texttt{sgoldt@sissa.it}}}

\affil[1]{Pritzker School of Molecular Engineering,
  University of Chicago, USA}
\affil[2]{ICTP -- The Abdus Salam International Centre  for Theoretical Physics,
  Trieste, Italy}
\affil[3]{Max Planck Institute for Dynamics and  Self-Organization, G\"ottingen, Germany}
\affil[4]{International School of Advanced Studies (SISSA),
  Trieste, Italy}

\date{}

\begin{document}

\maketitle
\begin{abstract}
  \input{abstract}
\end{abstract}

\input{main_text}

\section*{Acknowledgements}

\input{acknowledgements}

\printbibliography

\newpage
\appendix
\onecolumn
\numberwithin{equation}{section}

\input{appendix}

\end{document}

%% file: settings.tex
\newcommand{\bC}{\mathbf{C}}
\newcommand{\bX}{\mathbf{X}}
\newcommand{\bc}{\mathbf{c}}
\newcommand{\bh}{\mathbf{h}}
\newcommand{\bx}{\mathbf{x}}
\newcommand{\by}{\mathbf{y}}
\newcommand{\Sa}{\mathrm{Sa}}
\newcommand{\reals}{\mathbb{R}}

\definecolor{C0}{HTML}{1f77b4}
\definecolor{C1}{HTML}{ff7f0e}
\definecolor{C2}{HTML}{2ca02c}
\definecolor{C3}{HTML}{d62728}
\definecolor{C4}{HTML}{9467bd}
\definecolor{C5}{HTML}{8c564b}
\definecolor{C6}{HTML}{e377c2}
\definecolor{C7}{HTML}{7f7f7f}
\definecolor{C8}{HTML}{bcbd22}
\definecolor{C9}{HTML}{17becf}

%% file: abstract.tex
The recent success of neural networks in natural language processing has drawn
renewed attention to learning sequence-to-sequence (seq2seq) tasks. While there
exists a rich literature that studies classification and regression tasks using
solvable models of neural networks, seq2seq tasks have not yet been studied from
this perspective. Here, we propose a simple model for a seq2seq task that has
the advantage of providing explicit control over the degree of memory, or
non-Markovianity, in the sequences -- the stochastic
switching-Ornstein-Uhlenbeck (SSOU) model. We introduce a measure of
non-Markovianity to quantify the amount of memory in the sequences. For a
minimal auto-regressive (AR) learning model trained on this task, we identify
two learning regimes corresponding to distinct phases in the stationary state of
the SSOU process. These phases emerge from the interplay between two different
time scales that govern the sequence statistics. Moreover, we observe that while
increasing the integration window of the AR model always improves performance,
albeit with diminishing returns, increasing the non-Markovianity of the input
sequences can improve or degrade its performance. Finally, we perform
experiments with recurrent and convolutional neural networks that show that our
observations carry over to more complicated neural network architectures.

%% file: main_text.tex
\section{Introduction}


The recent success of neural networks on problems in natural language processing~\cite{devlin2018bert,
  howard2018universal, radford2018improving, brown2020language, chatgpt} has rekindled interest in tasks that require transforming a
sequence of inputs into another sequence (seq2seq). These problems also appear in many branches of science, often in the form of time series analysis~\cite{kantz2004nonlinear, box2015time}. 
Despite their ubiquity in science, there has been little work analysing learning of seq2seq tasks from a theoretical point of view in simple toy models. Meanwhile, a large body of work emanating from the statistical physics community~\cite{gardner1989three, seung1992statistical, engel2001statistical, mezard2009information, carleo2019machine} has studied the performance of simple neural networks on toy problems to develop a theory of supervised learning, where a high-dimensional input such as an image is mapped into a low-dimensional label, like its class.


An important recent insight from the study of supervised learning was the importance of data structure for the success of learning. The challenge of explaining the succes of neural networks on computer vision  tasks~\cite{krizhevsky2012imagenet, simonyan2015very, he2016deep,
  dosovitskiy2021image} inspired a new generation of data models that take
the effective low-dimensionality of images~\cite{pope2021intrinsic} into
account, such as object manifolds~\cite{chung2018classification}, the hidden
manifold~\cite{goldt2020modelling, goldt2022gaussian}, spiked
covariates~\cite{ghorbani2020neural, richards21asymptotics}, or low-dimensional
mixture models embedded in high dimensions~\cite{chizat2020implicit,
  refinetti2021classifying, loureiro2021learning}. 
By deriving learning curves for neural networks on these and other data
models~\cite{spigler2020asymptotic, dascoli2021interplay, benna2021place,
  gerace2022probing}, the importance of data structure for the success of neural
networks compared to other machine learning methods was clarified~\cite{ghorbani2019limitations, ghorbani2020neural, chizat2020implicit,
  refinetti2021classifying}. 

\begin{figure*}[t!]
  \centering
  \includegraphics[width=\linewidth]{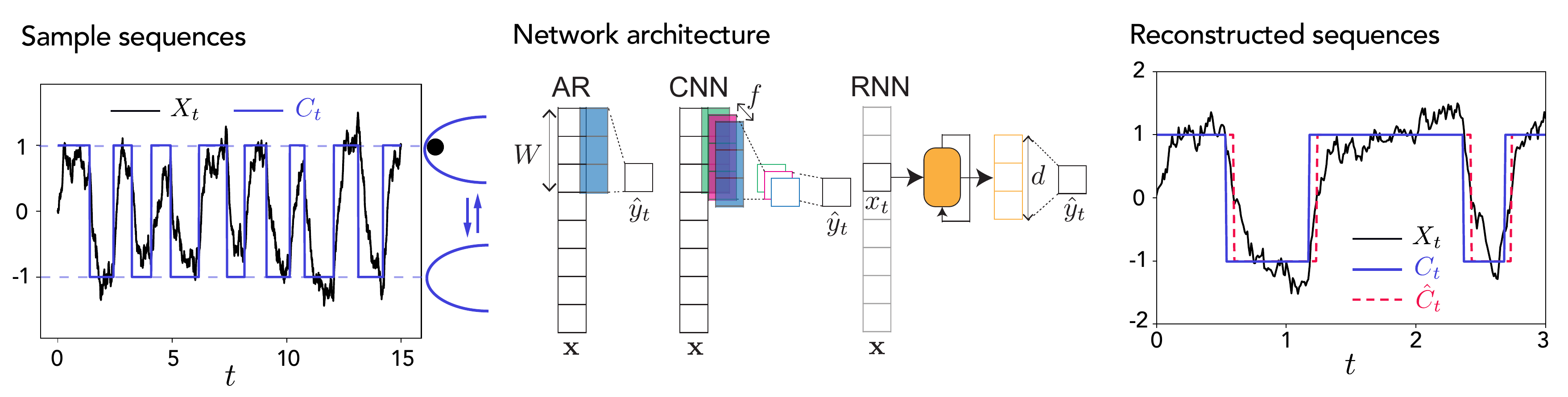}
  \vspace*{-1em}
  \caption{\label{fig:dynamics} \textbf{A flexible, minimal model for
      sequence-to-sequence learning tasks with varying degrees of memory along
      the sequence.}  \textbf{Left:} The motion of a Brownian particle (black
    filled circle) in a switching parabolic potential ``trap'' yields dynamics
    as given by the SSOU (stochastic switching Ornstein Uhlenbeck) model given
    by \cref{eq:eom-x}.  Example trajectories for sequences of the particle's
    position $X_t$ (black line) and the trap center $C_t$ (blue line) as shown
    as a function of time $t$. The blue dashed line is set at the trap
    centers~$C_0=\pm 1$. \textbf{Middle}: We train three types of models to
    reconstruct the trap positions $C_t$ from the particle trajectories $X_t$:
    auto-regressive models (AR) that makes predictions of the trap position based on the past $W$ observations of the particles position,    as well as 1D convolutional neural networks (CNN) that acts as a set of $f$ parallel AR models with added nonlinearity in the output, and recurrent neural
    networks (RNN) that use  feedback loops with a $d$ dimensional internal state to capture dependencies across time steps.\textbf{Right:} Sample reconstruction of the
    particle trap positions $\hat{y}_t$ (red dashed line) for one example input
    sequence ($X_t$, black solid line) compared with the actual hidden trap
    position ($C_t$, blue solid line). The sequence is a zoomed-in view of the
    sample sequence in the left panel.  \emph{Parameters}: $\kappa =10$, $k=1$,
    $D=0.5$, simulation time step $\Delta t=0.02$, AR model window size $W=2$.
    For this example, we used 5000 training samples, evaluated them over 5000
    test samples, a mini-batch size 32 and 5 epochs.}
    \vspace*{-1em}
\end{figure*}


The aforementioned breakthroughs on seq2seq tasks make
it an important challenge to extend the approach of studying learning in toy models to cover the realm seq2seq tasks. What are
key properties of sequences, akin to the low intrinsic dimension of images, that
need to be modelled?  How do these properties interact with different machine learning methods like auto-regressive models and neural networks?


Here, we make a first step in this direction by proposing to use a minimal, solvable latent variable model for time-series data that we call the stochastic switching Ornstein-Uhlenbeck process (SSOU). The input data consists of a sequence $\bx=(x_t)_{t=0}^T$, which depends on a latent, unobserved stochastic process $\bc = (c_t)_{t=0}^T$. The learning task is to reconstruct the unobserved sequence $\bc$ from the input sequence $\bx$, cf.~\cref{fig:dynamics}. The key data property we aim to describe and control is the \emph{memory} within the input sequence $\bx$. In the simplest scenario, the sequence~$x_t$ is memoryless: given the value of the present token $x_t$, the value of the next token $x_{t+1}$  is statistically dependent solely on  $x_t$ but not on previous tokens~$x_{t'<t}$.
Such a sequence can be described by a Markov process. By tuning the dynamics of
the latent process $c_t$, we can control the memory of the sequence,
\emph{i.e.}~we can increase the statistical dependence of future tokens on their
full history $x_{t'<t}$ (we introduce a precise, quantitative measure for
the memory below). Adding memory thus makes the process non-Markovian and allows
us to model richer statistical dependencies between tokens. At the same time,
the presence of memory in a process makes the mathematical analysis generally
harder. Our goal is to analyse how different neural network architectures --
auto-regressive (AR), convolutional (CNN), and recurrent (RNN) -- handle the
memory when solving the seq2seq task~$\bx \to \bc$. 

\noindent Our \textbf{main contributions} can be summarised as follows:
\begin{enumerate}
\item We introduce the stochastic switching Ornstein-Uhlenbeck (SSOU) process as
  a latent variable model for seq2seq tasks (\cref{sec:datamodel})
\item We introduce a measure of non-Markovianity that quantifies the memory of
  the sequences, and we describe how to tune the memory of the input and target
  sequences (\cref{sec:quantifying-non-markiovanity})
\item We use analytical~\cite{tucci2022modeling} and numerical evidence to identify two
  regimes in the performance of the AR model in the Markovian and non-Markovian case, respectively. These regimes   emerge from the interplay of two
  different time scales that govern the sequence statistics (\cref{sec:ar2})
\item We show that the task difficulty is non-monotonic in the sequence memory
  for all three architectures: the task is easy with strong memory or no memory,
  and hardest when there is a weak amount of memory. We explain this effect
  using the sequence statistics (\cref{sec:interplay_memory})
\item We finally find that increasing the integration window of auto-regressive
  models and the kernel size of CNN improves their performance, while increasing
  the dimension of the hidden state of gated RNN achieves only minimal
  improvements (\cref{sec:interplay_memory})
\end{enumerate}

\paragraph{Reproducibility} We provide code to sample from the SSOU model and
reproduce our experiments at
\href{https://github.com/anon/nonmarkovian_learning}{https://github.com/anon/nonmarkovian\_learning}.

\section{A model for sequence-to-sequence tasks}%
\label{sec:datamodel}

We first describe a latent variable model for seq2seq tasks, which we call the
Stochastic Switching-Ornstein-Uhlenbeck (SSOU) model. This model was introduced
recently in biophysics to describe experimental recordings of the spontaneous oscillations of the tip of
hair-cell bundles in the ear of the bullfrog~\cite{tucci2022modeling}. Furthermore, a simplified variant of the model with exponential waiting times has also been  fruitfully explored in the experimental context to describe the relaxation oscillations of colloidal particles in switching optical traps~\cite{pietzonka2017finite,di2023variance}.

We consider observable sequences $\bX = (X_t)$ which are described by a
one-dimensional stochastic process whose dynamics is driven by an autonomous
latent stochastic process $\bC = (C_t)$ and a Gaussian white noise that is
independent to $C_t$.  Here and in the following we index sequences by a time
variable~$t\ge 0$, and use bold letters such as $\bX$ to denote sequences, or
trajectories. A sequence of length $T$ can be sampled from the stochastic
differential equation
\begin{equation}
  \label{eq:eom-x}
  X_{t+\mathrm{d} t} - X_{t} = - \kappa (X_t - C_t) \mathrm{d} t + \sqrt{2D}\mathrm{d} B_t,
\end{equation}
with $t\in [0,T-1]$. Here,  $\mathrm{d} B_t$ is the increment of the Wiener process in
$[t,t+\mathrm{d} t]$, with $\langle \mathrm{d} B_t \rangle=0$ and $\left\langle (\mathrm{d} B_t) (\mathrm{d}
  B_{t'}) \right\rangle=\delta(t-t')\mathrm{d} t$. The angled brackets $\langle \cdot
\rangle$ indicate an average over the noise process. We denote the parameters
$D>0$ and $\kappa>0$  as diffusion coefficient and trap stiffness for reasons
described below.

We employ this setup as a seq2seq learning task where we aim to reconstruct the
hidden sequence of trap positions~$\bC$ given a sequence of particle positions
$\bX$, see~\cref{fig:dynamics} for an illustration. The key idea in our model is
to let the location of the potential $C_t$ alternate in a stochastic manner
between the two positions $C_0=\{-1, 1\}$. The waiting time $\tau$ spent in each
of these two positions is drawn from the waiting-time distribution
$\psi_k(\tau)$. For a generic choice of $\psi_k(\tau)$, the process $C_t$ -- and
hence $X_t$ -- is non-Markovian and has a memory. Here, we use a one-parameter
family of gamma distributions (defined in \cref{eq:waitingtimes} below), which
allows us to quantitatively control the degree of memory in the sequence of
tokens $\bC$ in a simple manner; we discuss this in detail in
\cref{sec:quantifying-non-markiovanity}.

\subsection{Physical interpretation of the SSOU model}

A well-known application of \cref{eq:eom-x} in statistical physics is to
describe the trajectories of a small particle (e.g. a colloid) in an aqueous
solution undergoing Brownian motion~\cite{vankampen1992stochastic} in a
parabolic potential with time-dependent center $C_t$. Such physical model is
often realized with microscopic particles immersed in water and trapped with
optical
tweezers~\cite{martinez2012force,pietzonka2017finite,martinez2013effective}. When
$\kappa=0$, the particle is driven only by the noise term $\mathrm{d} W_t$ and
performs a one-dimensional free diffusion along the real line with diffusion
coefficient $D$. By choosing $\kappa$ positive, the particle experiences a
restoring force $-\kappa (X_t - C_t)$ towards the instantaneous center of the
potential. Such force
$F(X_t,C_t) = -\left.\partial_X V(X,C)\right|_{X=X_t,\,C=C_t}$ tends to confine
the particle to the vicinity of the point $C_t$, and is therefore often called a
particle trap, which is modelled by a harmonic potential centred on $C_t$
\begin{equation}
  \label{eq:potential}
  V(X_t, C_t) = \frac{\kappa}{2}{(X_t-C_t)}^2.
\end{equation}
This motivates the name ``stiffness'' for $\kappa$; the higher $\kappa$, the
stronger the restoring force that confines the particle to the origin of the
trap $C_t$. From a physical point of view, the dynamics of $X_t$ consists of the
alternate relaxation towards the two minima of the potential between consecutive
switches, cf.~\cref{fig:dynamics}.


\section{Architectures and training}%
\label{sec:methods}



\paragraph{Auto-regressive model} We first consider  the arguably simplest machine-learning model that can be
trained on sequential data, an auto-regressive model of order $W$,  AR($W$), see \cref{fig:dynamics}. The output $\hat \by =
(\hat y_t)$ of the model for the trap position given the sequence $\bx$ is given by
\begin{equation}
    \label{eq:ar-model}
    \hat y_t = \sigma\left(\sum_{\tau=1}^W w_\tau x_{t-\tau+1}+b\right)
\end{equation}
where $\sigma(x)=1/(1+e^{-x})$ is the sigmoidal activation function, and the
weights $w_\tau$ and the bias $b$ are trained. Its basic structure -- one layer
of weights followed by a non-linear activation function -- makes the AR model
the seq2seq analogue of the famous perceptron that has been the object of a
large literature in the theory of neural networks focused on
classification~\cite{gardner1989three, seung1992statistical,
  engel2001statistical, carleo2019machine}.  Note that the window size $W\ge 1$
governs the number of tokens accessible to the model.  We do not pad the input
sequence, so the output sequence is shorter than the input sequence by $W-1$
steps.

\paragraph{Convolutional two-layer network} 
The auto-regressive model can also be thought of as a single layer 1D
convolutional neural network~\cite{lecun1989backprop, goodfellow2016deep} with a
single filter and a kernel of size $W$ with sigmoid activation. We also consider
the natural extension of this model, CNN($W$), a 1D convolutional neural network
with two layers (see \cref{fig:dynamics}). The first layer has the same kernel
size ($W$), but contains $f$ filters with rectified linear
activation~\cite{fukushima1969visual}. In the second layer, we apply another
convolution with a single filter and a kernel size of 1 with sigmoid
activation. In this way, we can compare a CNN and an AR model whose
``integration window'' are of the same length $W$. This allows us to investigate
the effect of additional nonlinearities of the second layer of the CNN on the
prediction error.

\paragraph{Recurrent neural network} We also apply a recurrent network to this task that takes $x_t$ as the input at step $t$, followed by a layer of Gated Recurrent Units~\cite{cho2014properties} with a $d$-dimensional hidden state, followed by a fully connected layer with a single output and the sigmoid activation function (see \cref{fig:dynamics}). We refer to this family of models by GRU($d$).

\paragraph{Generating the data set}  We train all the models on a training set with $N$ sequences
$\{ \bx^{(n)} \}_{n=1}^N$, which can in principle be of different lengths. To obtain a
sequence, we first generate a trajectory  $\bC$ 
for the latent variable by choosing an initial condition at random from $C_0 \in \{-1, 1\}$ and then drawing
a sequence of waiting times from the distribution $\psi_k(\tau)$. Using the
sampled trap positions, $\bC$, 
we then sample the trajectory $\bX$ from \cref{eq:eom-x}. Since in practice we do not have access to the full trajectory, as any experiment has a finite sampling rate, we subsample the process $X_t$ by
taking every $s$th element of the sequence. That is, for a given sequence
$\{ X_t' \}_{t'=0}^{T'}$ of length $T'$, we construct a new sequence
$\bx={(x_{t})}_{t=1}^{T}$, where $x_{t} = X_{s\times t}$ and
$T = \lfloor T' / s \rfloor$. The subsampled sequence $\bx$ is then used as an
input sequence. We verified that the finite time step for the integration of the SDE and the subsampling preserve the statistics of the continuum description that we use to derive our theoretical results, cf.~\cref{app:subsampling}. For convenience, we also introduce $\by=(1+\bc)/2$ to shift the trap position to 0 or 1. We then use this subsampled and shifted sequence of trap positions as the target.  

\paragraph{Training and evaluation} We train the models by minimising the mean
squared loss function using Adam~\cite{kingma2014adam} with standard hyperparameters. For each sample sequence, the loss is defined as
\begin{equation}
    \label{eq:loss}
    \ell(\by_t, \hat \by_t) = \frac{1}{T-\tau_0}\sum_{\tau=\tau_0}^T (y_\tau - \hat y_\tau)^2,
\end{equation}
where the offset $\tau_0$ in the lower limit of the sum is necessary for the AR and CNN models since
their output has a different length than the input
sequence. For those models we choose $\tau_0 = W-1$. For RNN models, however, we use $\tau_0=1$ as the input and out sequence lengths are the same. 
We assess the performance of the trained models by first thresholding their
outputs $\hat{y}_t$ at 0.5 to obtain the sequence~$\hat{c}_t \in \{\pm 1\}$,
since the location of the centre of the potential has two possible values $\pm 1$. We then use the misclassification error $\epsilon$ defined as
\begin{equation}
    \epsilon = \frac{1}{N_{\rm{samples}}N_\tau}\sum_{{\rm{samples}},\tau>\tau_h}\delta(c_\tau\neq \hat{c}_\tau),
\end{equation}
where $N_\tau$ is the length of the test sequence and $\delta(x)=1$ if the condition $x$ is true and it is 0 otherwise, as the figure of merit throughout this work. Unless otherwise specified, we used 50000 training samples with mini-batch size 32 to train the models, and
we evaluated them over 10000 test samples. To consistently compare different models with different output lengths and to reduce the boundary effects we only consider the predictions after an initial time offset~$\tau_h$. 


\section{Results}

\subsection{The waiting time distribution of the trap controls the memory of the
  sequence}%
\label{sec:quantifying-non-markiovanity}

\begin{figure*}[t!]
  \centering
  \includegraphics[width=0.45\linewidth]{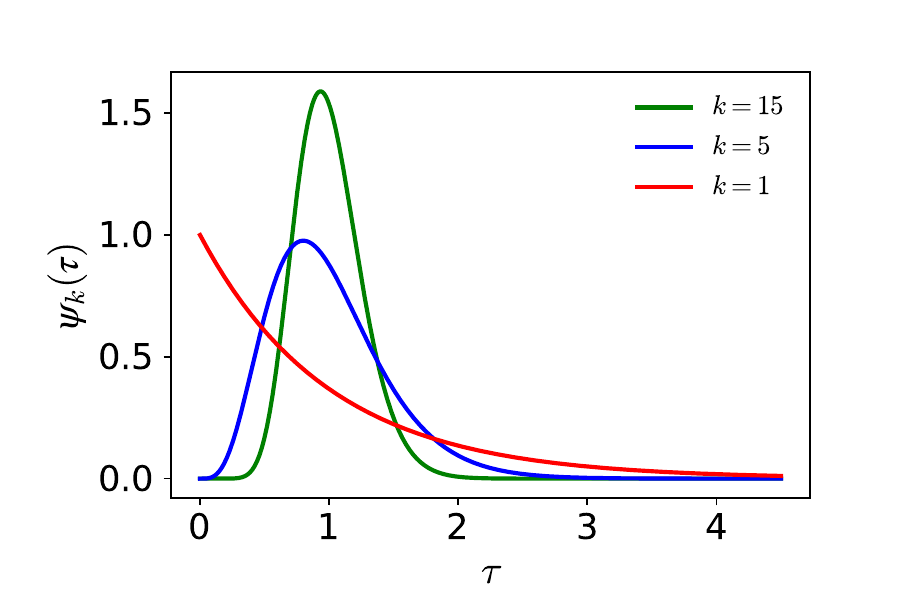}\quad%
  \includegraphics[width=.45\linewidth]{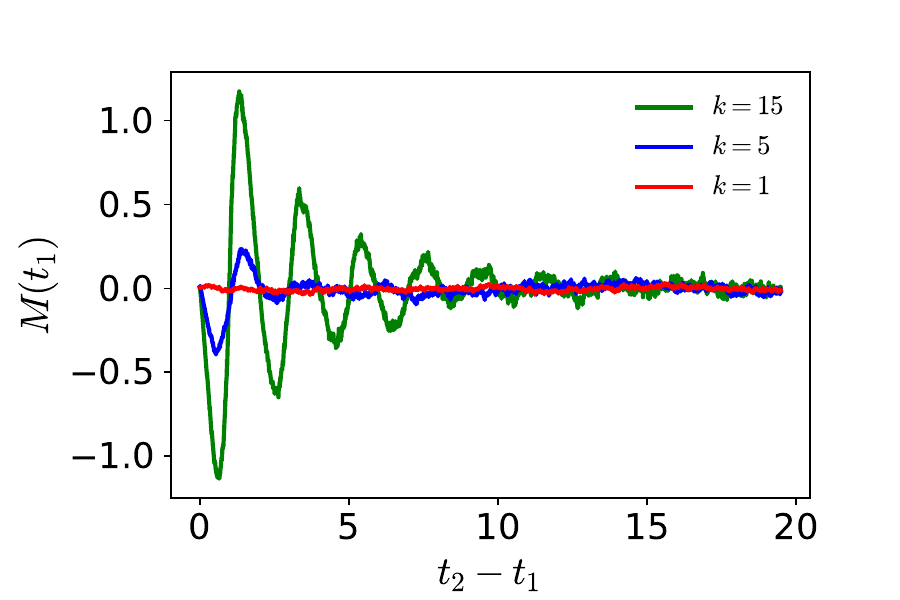}
  \caption{\label{fig:non-markovianity} \textbf{Quantifying the memory of
      non-Markovian input sequences.} Left: waiting-time
    distribution~$\psi_k(\tau)$ given by \cref{eq:waitingtimes} for three
    choices of $k$. Right: measure~${M}(t)$ (see \cref{eq:non-markovianity})
    associated with the parameters in the left panel, which quantifies the
    memory of the past time $t_1$ in the sequence. We obtain a memoryless
    sequence $X_t$ with ${M}(t)=0$ by choosing ${k}=1$ which creates an
    exponential waiting-time distribution, \cref{eq:waitingtimes} (red
    curves). As the value of $k$ increases, the memory becomes stronger (blue
    and green curves).
    \emph{Parameters}: 
    $D=0.5$, $\kappa=2$, $\Omega_1=\Omega_2=[0.5,1.5]$, $t_3-t_2=0.5$, by
    generating $N=10^5$ trajectories via the Euler–Maruyama
    method~\cite{kloeden1992stochastic} with simulation time step
    $\Delta t=5\times 10^{-3}$.  }
\end{figure*}

The key data property we would like to control is the memory of the sequence $\bX$. In our SSOU model, the memory is controlled 
by tuning the memory in the latent sequence $\bC$, which in turn depends on the distribution of the  waiting time~$\tau$. To fix ideas, we choose gamma-distributed waiting
times,
\begin{equation}
    \label{eq:waitingtimes}
    \psi_k(\tau)=\frac{1}{\tau}\frac{(\tau k)^k e^{-k\tau}}{\Gamma({k})},
\end{equation}
where we set ${k}\ge 1$, while $\Gamma(k)=\int_0^{\infty} \mathrm{d} x\, x^{k-1} e^{-x} $ denotes the
Gamma function. Note that for any choice of $k$, the normalization condition is  $\int_0^\infty\psi_k(\tau)\mathrm{d} \tau=1$ and the average waiting time between two consecutive switches is $\langle\tau\rangle_k=\int_0^\infty \tau\psi_k(\tau)\,\mathrm{d}\tau=1$. We can control the shape of the distribution by changing the value of $k$: the variance of the waiting time for example is $k$-dependent, $\langle \tau^2\rangle_k  -\langle \tau\rangle_k^2 = 1/k$ (cf.~\cref{fig:non-markovianity}). Furthermore, $k$ also controls the ``degree of non-Markovianity'' of the latent sequence $\bC$ as follows.

For the  choice ${k}=1$, the waiting-time distribution
is exponential, making the process $C_t$ Markovian. 
This result, together with the fact that \cref{eq:eom-x} is linear, 
implies that the observable process $X_t$ is Markovian, too. Thus 
the  probability distribution for the tokens $x_t$  obeys the
Markov identity~\cite{vankampen1992stochastic}
\begin{equation}
  \label{eq:markovianity}
  p_{1|2}(x_{t_3} | x_{t_2}; x_{t_1}) = p_{1|1}(x_{t_3} | x_{t_2}) ,
\end{equation}
for any $t_1<t_2<t_3$, which represent different instants in time, and $p_{1|m}$ denotes the conditional probability density (at one time instant) with $m$ conditions (at $m$ previous time instants).
\Cref{eq:markovianity} is the mathematical expression of the intuitive argument that we gave earlier: in
a Markovian process, the future state  $x_{t_3}$ at some time $t_3$ given the state $x_2$ at some earlier time $t_2<t_3$ is conditionally independent of the
previous tokens $x_1$ for all times $t_1<t_2$. 

However, most of the sequences analysed in machine learning are non-Markovian:
in written language for example, the next word does not just depend on the
previous word, but instead on a large number of preceding words. We can generate
non-Markovian sequences by choosing ${k} >
1$. 
To systematically investigate the impact of sequence memory on the learning
task, it is crucial to quantify the \textit{degree} of non-Markovianity of the
sequence for a given value of $k$ beyond the binary distinction between
Markovian and non-Markovian. Yet defining a practical measure that quantifies
conclusively the degree of non-Markovianity is a non-trivial
task~\cite{lapolla2021toolbox, lapolla2019manifestations} and subject of ongoing
research mainly done in the field of open quantum systems
\cite{laine2010measure, hall2014canonical, rivas2010entanglement,
  huang2021quantifying, strasberg2018response}.


Here, 
we introduce a simple quantitative measure of the degree of non-Markovianity of
the input sequence motivated directly by the defining property of Markovianity given in
\cref{eq:markovianity}, which reads
%
%
\begin{equation}
  \label{eq:non-markovianity}
  \begin{split}
  M(t_1,t_2,t_3)
  &\equiv \frac{\left \langle x_{t_3} |~ x_{t_2}\!\in {\Omega_2}, x_{t_1}\!\in
      {\Omega_1}\right\rangle }{\left\langle x_{t_3} |~ x_{t_2}\in
      {\Omega_2} \right\rangle} - 1 \\
  & = \frac{\int_{\mathcal{X}} x_{t_3}\; p_{1|2}(x_{t_3} |~ x_{t_2}\!\in
    {\Omega_2}, x_{t_1}\!\in {\Omega_1}) \mathrm{d} x_{t_3}}{\int_{\mathcal{X}}
    x_{t_3}\; p_{1|1}(x_{t_3} |~ x_{t_2}\!\in {\Omega_2}) \mathrm{d} x_{t_3}} - 1.
\end{split}
\end{equation}
\Cref{eq:non-markovianity} involves crucially conditional expectations\footnote{We denote $\langle X | Y\in \Omega\rangle$  the conditional expectation of the random variable $X$ given that the random variable $Y$ satisfies a certain criterion, symbolyzed here as belonging to the set $\Omega$. Note that in general $X$ and $Y$ are statistically dependent, and that the conditioning  may be done with respect to more than one random variables satisfying prescribed criteria. }, where the expectation of the future system state (at time $t_3$) is conditioned on the present state (at time $t_2$), or on the present and past state (at times~$t_2$ and~$t_1$). 
We have further introduced the notion of state space regions $\Omega_1\subseteq \mathcal{X}$ and $\Omega_2\subseteq \mathcal{X}$ which are subsets of the entire state space $\mathcal{X}= \mathbb{R}$ that is accessible by the process $\bX$. 
The measure ${M}$ defined in \cref{eq:non-markovianity} quantifies the drop in uncertainty 
about the future mean values, when knowledge about the past ($x_{t_1}$) is given in addition to the knowledge of the present state. Because of noise, this generally depends on how far from the past the additional data point is, $t_2-t_1$, where the decay of ${M}$ with $t_2-t_1$ quantifies the ``memory decay'' with increasing elapsed time.
From~\cref{eq:markovianity} it directly follows that 
 for any Markovian process,
the measure 
${M}$ defined in \cref{eq:non-markovianity} vanishes  at all times, whereas ${M}\neq 0$ reveals the presence of non-Markovianity in the form of memory of the past time $t_1$. 
%
In a stationary process,~$M$ generally depends on the two time differences $t_3-t_2$ and $t_2-t_1$, and on the choices of $\Omega_1$ and $\Omega_2$. To spot the non-Markovianity, $t_3-t_2$ should be comparable to (or smaller than) the dynamical time scales of the process, which can be defined by the decay of correlation functions, which we show in fig.~\ref{fig:Cx} in the appendix (namely, if $t_3-t_2$ is so large that $x_{t_3}$ and $x_{t_2}$ are fully uncorrelated, $M$ trivially vanishes even for non-Markovian processes). The choice of $\Omega_1$ and $\Omega_2$ is in principle arbitrary, but, for practical purposes, they should correspond to regions in the state space  that are frequently visited.

We plot ${M}$ in \cref{fig:non-markovianity} for three different values of ${k}$ obtained from numerical simulations of the SSOU. Here, we fix $t_{2,3},\Omega_{1,2}$, and vary $t_1$. As expected, for a Markovian switching
process with ${k}=1$, ${M}(t_1)$ vanishes at all times $t_1$, while for ${k}>1$, ${M}(t_1)$ displays non-zero values. The non-Markovianity measure $M(t_1)$ defined in \cref{eq:non-markovianity} generally captures different facets of non-Markovianity. On the one hand, the decay with $t_2-t_1$ measures how far the memory reaches into the past. On the other hand, the magnitude of $M(t_1)$ tells us how much the predictability of the future given  the present state profits from additionally knowing the past state at time $t_1$.
For the SSOU model, we observe in \cref{fig:non-markovianity} that increasing ${k}$ increases both the decay time and the amplitude of ${M}$, showing that that the parameter ${k}$ controls conclusively the degree of non-Markovianity. 
We further note that ${M}(t_1)$ displays oscillations for ${k}>1$, reflecting the oscillatory behaviour of $\bC$ and~$\bX$, and that  ${M}(t_1)$ always decays to zero for sufficiently large values of
$t_2 - t_1$, indicating the finite persistence time of the memory in the system. 
We note that these observations are robust with respect to the details of the non-Markovianity measure. For example, we numerically confirmed that when we change $t_3-t_2$, or $\Omega_{1,2}$, the essential features of $M$ and its dependence on $k$ remain unchanged.

In conclusion of this analysis, we can in the following simply use ${k}$ as control parameter of the degree of non-Markovianity of $\bX$.

\subsection{The performance of auto-regressive models}%
\label{sec:ar2}

\subsubsection{The interplay between two time scales determines prediction error}

\begin{figure*}[t!]
  \centering
  \includegraphics[width=.9\linewidth]{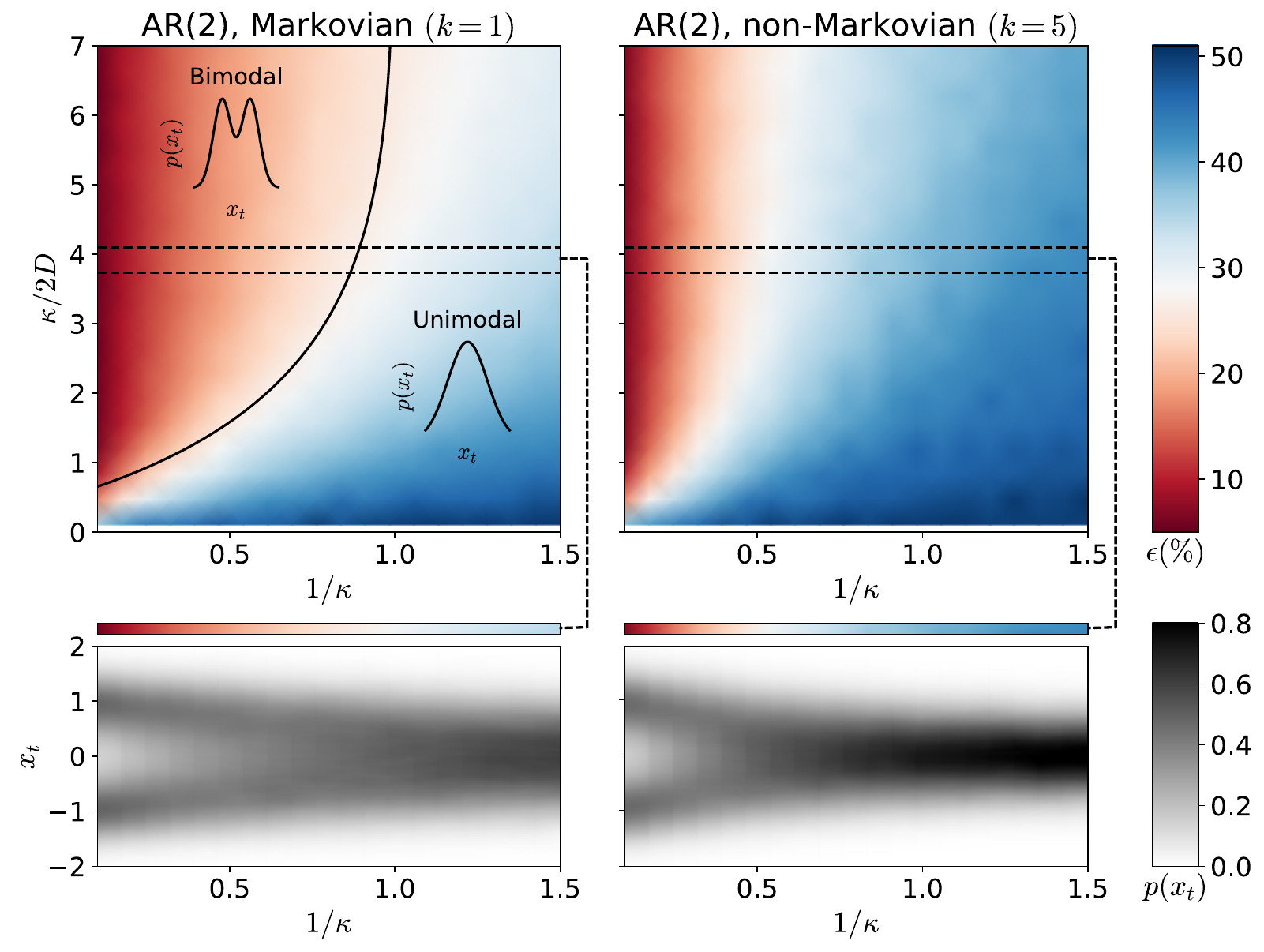}
  \caption{\label{fig:eq-phase-diagram} \textbf{Performance of auto-regressive
      AR(2) models for Markovian and non-Markovian sequences of the SSOU model
      across its ``learnability phase diagram''.}  
    Reconstruction error of the AR(2) model (in \% of correctly predicted trap positions) for different values of
    the diffusive time scale $t_{\rm diff} = 1/2D$ and the relaxation time
    $t_\kappa = 1 / \kappa$, see \cref{eq:time scales}.  We show these ``phase
    diagrams'' for \textbf{(Left)} Markovian sequences ($k=1$) and for \textbf{(Right)} non-Markovian sequences
    ($k=5$). In these diagrams, the two axes represent ratios of time scales,
    $t_{\rm diff}/t_\kappa$ (vertical axis) vs
    $t_\kappa/\langle \tau \rangle_k= 1/\kappa$ (horizontal axis), since
    $\langle \tau \rangle_k=1$ for all
    $k$.  The bottom row depicts the distribution of particle's position $p(x_t)$ across the dashed line superimposed with the error heat map. The density shows a clear transition from a bimodal distribution to a unimodal one. 
    \emph{Parameters:} total simulation time for each parameter value $\tau=30$,
    simulation time step $\Delta t$=0.01, AR($W$) window size $W=2$, $\kappa$ varying
    from 0.1 to 0.6, $\tau_h=2$, remaining training parameters as in~\cref{fig:dynamics}. 
    }
\end{figure*}

To gain some intuition, we first consider the simplest possible students, namely
auto-regressive models AR(2) from \cref{eq:ar-model} with window size $W=2$.  The student predicts the value of $c_t$ only using information about the present and the previous tokens $x_t$ and $x_{t-1}$, giving it access to the current particle position and allowing it in principle also to estimate the velocity of the particle. We show the performance of this model obtained numerically in \cref{fig:eq-phase-diagram}. The error of AR(2) varies significantly from less than $5\%$ to essentially random guessing at~$50\%$ error as we vary the two time scales that influence the sequence statistics,
\begin{equation}
    \label{eq:time scales}
    t_\kappa=1/\kappa \qquad \mathrm{and} \qquad t_{\rm diff} = C_0^2/2D=1/2D
\end{equation}
the relaxation time and the diffusive time scale. The \textbf{relaxation
time}~$t_\kappa$ determines how quickly
the average position of the particle is restored to the centre of the trap when the the average waiting time $\langle\tau\rangle_k=1$. A faster relaxation time means that the particle responds more quickly to a change in~$c_t$, making reconstruction easier. The \textbf{diffusive time scale} $t_{\rm diff}$ sets the typical time that the system elapses to randomly diffuse a distance $\sim |C_0|$. The larger this time scale, the slower the particle moves randomly, as opposed to movement that follows the particle trap, hence improving the error.



\subsubsection{A ``phase diagram'' for learnability} 

For memoryless sequences ($k=1$), we can explain this performance more quantitatively by solving for the data distribution $p(x_t)$. Interestingly, we observe that along the ``phase boundary'' between unimodal and bimodal particle distribution $p(x_t)$, the error is roughly homogeneous and approximately equal to $25\%$.  The presence of two ``phases''  in terms of the bimodality/unimodality of the distribution $p(x_t)$ is rationalised using recent analytical results for the loci of the phase boundary in the the case $k=1$~\cite{tucci2022modeling}, which is given by
  \begin{equation}
\label{eq:chistar}
    \frac{t_{\rm diff}}{t_\kappa}=\frac{(t_\kappa+1/2)\,\,{}_1F_1\left(1/2,t_\kappa+1/2,-t_{\rm diff}/t_\kappa\right)}{{}_1F_1\left(3/2,t_\kappa+3/2,-t_{\rm diff}/t_\kappa\right)}.
 \end{equation}
This analytical result is obtained by finding the parameter values at which the
derivative of $p(x_t)$ at~$x_t=0$ changes sign, which corresponds to a
transition between unimodal and bimodal (see \cref{sec:app-phase-diagram} for
a detailed derivation). The line separating the two phases is drawn in black in the
density plot of \cref{fig:eq-phase-diagram}. In particular, 
we observe that reconstructing the trap position is simplified when the marginal distribution of the particle distribution~$p(x_t)$ is bimodal, i.e.~when it presents two well defined peaks around the trap centres $\pm C_0$   (see~\cref{fig:eq-phase-diagram} top left). On the contrary, when the distribution $p(x_t)$ is unimodal (corresponding to the case of  fast relaxation times  $t_\kappa$) the learning performance worsens as the  data is not sufficiently informative about the latent state $\bC$.  

If we add memory to the sequence by
choosing $k>1$, an analytical solution for the phase boundary is challenging, as it requires knowledge of the steady-state distribution of the joint system. For $k=1$, the system is Markovian and we can solve the Fokker-Planck equation for the steady-state distribution. For $k \neq 1$, the steady-state distribution is the solution of an integro-differential equation which is not solvable analytically. However, we can verify whether the same mechanism -- a transition from a unimodal to a bimodal distribution for the particle distribution $p(x_t)$ -- drives the error also in the non-Markovian case by means of numerical simulations. Interestingly, when increasing the degree of memory to $k=5$
(top right of \cref{fig:eq-phase-diagram}), we find that the AR(2)
model yields a larger error than for  $k=1$ Markovian sequences for all the
parameter values explored in the learnability phase diagram. This means that not
only that AR(2) is not able to extract all the available information from the
sequence, but that the task has become harder by increasing $k$. A look at the density plots for the particle distribution $p(x_t)$ explains this finding: the transition to unimodality happens for a smaller value of the relaxation time $t_\kappa$ in the non-Markovian case (bottom right of \cref{fig:eq-phase-diagram}). We go into detail on how the degree of non-Markovianity makes the task harder by quantifying this transition in the probability distribution in the next section.


\paragraph{Further statistical characterisation of the error}
\begin{figure}
    \centering
    \includegraphics[width=0.9\linewidth]{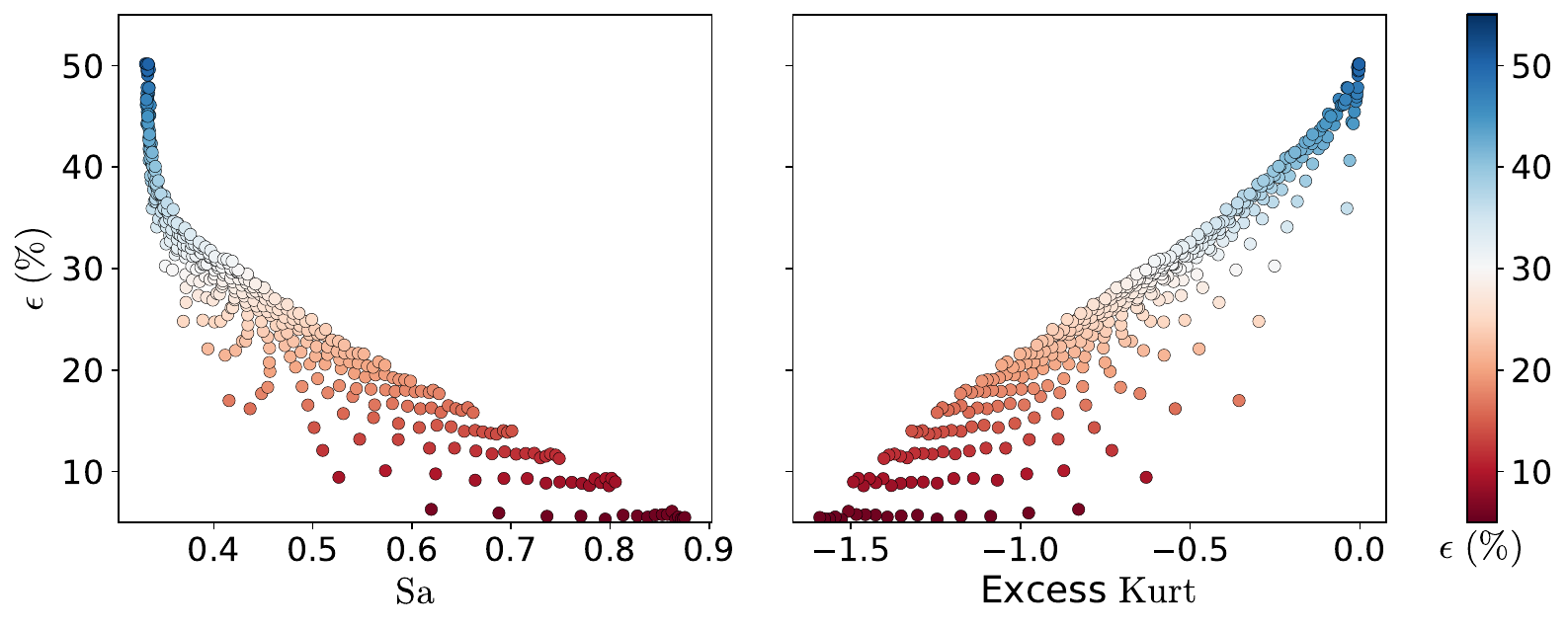}
    \caption{\label{fig:stats}\textbf{Statistical characterisation of the error} \textbf{(Left)} Scatter plot of the error of several AR(2) models trained
    on Markovian sequences versus the Sarle coefficient, a
    measure of the bimodality of the distribution $p(x_t)$, \cref{eq:sarle}.
    \textbf{(Right)} Scatter plot of the error of several AR(2) models trained
    on Markovian sequences versus the excess kurtosis, \cref{eq:excesskurt}, another
    measure of the bimodality of the distribution. \emph{Parameters}: Similar to those used in the right panel of \cref{fig:eq-phase-diagram}}
\end{figure}

We can quantify the degree of
bimodality of the particle distribution $p(x_t)$, which drives the error of the auto-regressive model, by using the Sarle coefficient~\cite{ellison1987effect},
\begin{equation}
    \label{eq:sarle}
    \Sa=(\sigma^2+1)/\mathcal{K}
\end{equation}
with $\sigma$ and $\mathcal{K}$ the skewness and kurtosis of $x_t$, respectively. The Sarle coefficient takes values from 0~to~1 and is equal to 5/9 for the uniform distribution. Higher values may indicate bimodality. We can see a clear correlation between the Sarle coefficient and the error ($\epsilon$) of the AR(2) model from the scatter plot on the right of \cref{fig:eq-phase-diagram}. Indeed, there appears to exist an upper bound on the error in terms for the Sarle coefficient. Another measure we can compute is the excess kurtosis, which for a zero-mean random variable reads
\begin{equation}
    \label{eq:excesskurt}
    \mathcal{K}_{\rm ex} = \frac{\langle x^4\rangle}{\left(\langle x^2\rangle\right)^2} - 3,
\end{equation}
where the average is over $p(x_t)$ and we have used the fact that $\langle x_t \rangle=0$ by symmetry. The excess kurtosis vanishes for a Gaussian distribution and is positive or negative for non-Gaussian distributions, depending on the behaviour of the tails of the distribution (hence the name \emph{excess} kurtosis). A negative value of the excess kurtosis instead indicates that the distribution is sub-Gaussian or platykurtic, i.e. having narrower tails than the Gaussian distribution. In our case, we obtain negative values of excess kurtosis whose magnitude gets larger when the distribution looks more bimodal.  Furthermore, we find that the larger the more negative is the excess kurtosis, the lower is the prediction error of the AR(2) model (right of \cref{fig:stats}). Taken together, this analysis shows that the prediction error becomes large when  the peaks of the distribution become less distinguishable and/or in the presence of fat tails.


\subsection{The interplay between sequence memory and model architecture}%
\label{sec:interplay_memory}

\subsubsection{The statistical properties of the input sequence $\bx$ with memory}

\begin{figure*}[t!]
  \centering
  \includegraphics[width=.9\linewidth]{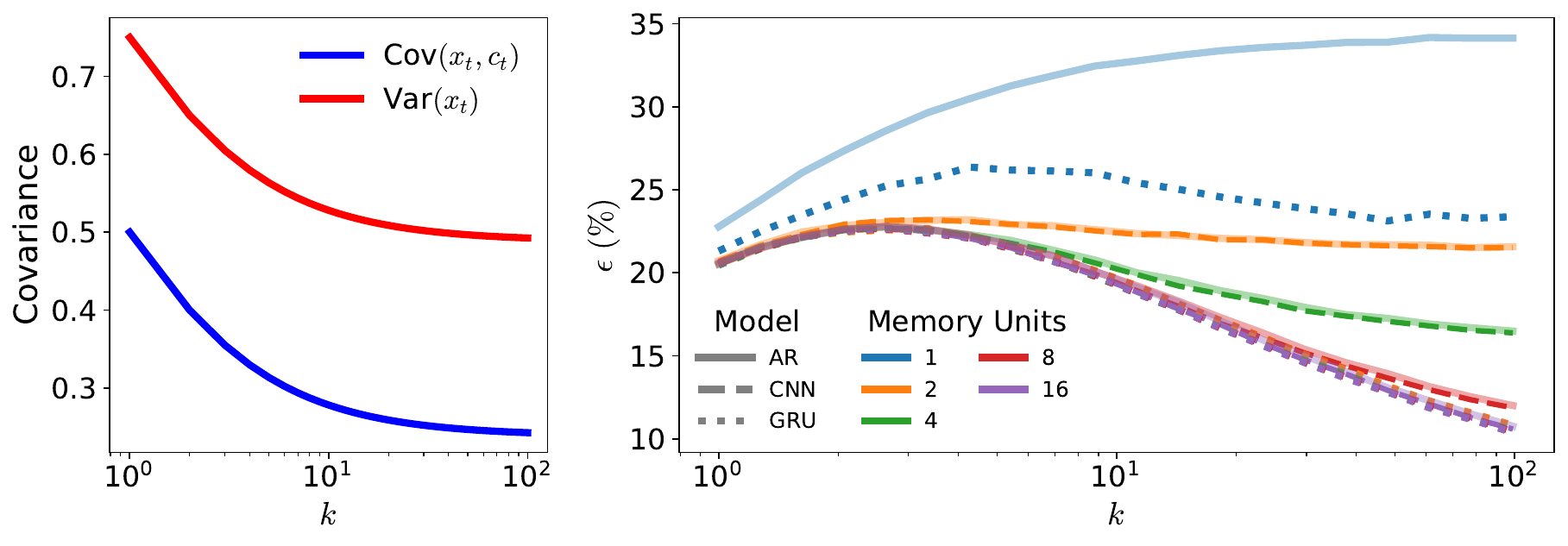}
  \caption{\label{fig:error-vs-k} \textbf{The impact of sequence memory on
      recurrent and convolutional neural networks.} \textbf{(Left)} Analytical
    predictions for the covariance of the particle and trap position,
    \cref{eq:correlations}, in the non-Markovian case as a function of $k$. As
    we increase the non-Markovianity, correlations between $x_t$ and $c_t$
    decay, complicating the reconstruction task. \textbf{(Right)} Prediction
    error~$\epsilon$ for students with different architectures: auto-regressive
    (AR), convolutional (CNN) and gated recurrent neural network
    (GRU). 
    The ``memory units'' of the architectures correspond to the size of the
    kernel ($W$) in the AR and the first layer of CNN models, and the number of
    units ($d$) in the hidden state of GRU models. The CNN models all have
    $f=10$ filters in their first layer. \emph{Parameters}: Similar to
    \cref{fig:eq-phase-diagram} with $\kappa=2$.}
\end{figure*}

We saw in \cref{fig:eq-phase-diagram} that increasing the sequence memory by
increasing the control parameter $k$ of the waiting time distribution,
\cref{eq:waitingtimes}, makes the problem \emph{harder}: the error of the
same student increased. We can understand the root of this difficulty by
studying the variance of the distribution~$p(x_t, c_t)$. Using the tools
introduced by \textcite{tucci2022modeling}, we can calculate these correlations
analytically for any integer $k$. As we describe in more detail in
\cref{sec:app-correlations}, we find that
\begin{subequations}
    \label{eq:correlations}
    \begin{align}
        \mathrm{Var}\; (x_t) &= \langle x_t^2 \rangle =\frac{D}{\kappa}\,+ C_0^2\kappa\sum_{n=0}^{{k}-1}\,\frac{Q_n}{{k}\,q_n+\kappa},\\
        \mathrm{Cov}\;( x_t, c_t) &= \langle x_t c_t \rangle =  1-\frac{2}{\kappa}\frac{(1+\kappa/k)^{k}-1}{(1+\kappa/k)^{k}+1},
    \end{align}
\end{subequations}
where $Q_n\equiv -4(1-q_n)/({k}\,q_n)^2$, and $q_n\equiv 1-e^{i\pi(1+2n)/{k}}$
with $n=0,\dots,{k}-1$; we recall
that~$ \langle x_t \rangle = \langle c_t\rangle = 0$. We plot both correlation
functions as we vary $k$ on the right of \cref{fig:error-vs-k}. First, we note
that the variance of the particle distribution decreases with $k$ (red line). In
other words, as we increase the memory of the sequence, the particle spends on
average more time around the origin, making reconstruction harder. This is also
born out by a decrease in the correlation between $x_t$ and $c_t$ (blue
line). This loss in correlations is closely related to the error of the simplest
reconstruction algorithm, where we estimate the particle positions $\bc$ by
thresholding the particle position, $c_t = \mathrm{sign}(x_t)$, and $|C_0|$ identifies the amplitude of the $C_t$ process. As shown by a
light blue line in the right panel of \cref{fig:error-vs-k}, the error of this
parameter-free, memoryless algorithm increases monotonically with $k$.

\subsubsection{Increasing the integration window of autoregressive and
  convolutional neural networks} The existence of memory in the sequences
suggests choosing student architectures that can exploit these additional
statistical dependencies. We first studied the performance of AR($W$) models
with varying~$W$. As shown in \cref{fig:error-vs-k}, we observed that for a
fixed $k$ increasing $W$ generally helps with reducing the error~$\epsilon$
(solid lines), so models with larger integration window can take advantage of
the additional temporal correlations in the sequence and make better predictions
in the large $k$ limit. For the convolutional networks described in
\cref{sec:methods} with $f=10$ filters (dashed lines), we observe that the
additional filters and non-linearities in the CNN do not help with the
predictions compared to the AR($W$) models. We conclude that it is truly the
size of the integration window that makes a difference on this data set and that
even a simple AR model can completely take advantage of the information in a
given time window. 

\subsubsection{Recurrent neural networks} We numerically studied the performance
of a gated recurrent unit GRU($d$)~\cite{cho2014properties} with hidden state
size $d$, which can be thought of as a slightly more streamlined variant of the
classic long short-term memory~\cite{hochreiter1997long}. Although such an RNN
reads in the input sequence one token at a time, it can exploit long-range
correlations by continuously updating its internal state $\bh \in \reals^d$,
cf.~\cref{fig:dynamics}. We show the test performance of GRU models with various
internal state sizes $d$ in \cref{fig:error-vs-k}. The GRUs display a good
performance: as soon as the GRU($d$) has a hidden state with dimension~$d=2$, it
captures all the memory of the sequence at all $k$, achieving the limiting
performance of AR($W$) models with the largest window size. Indeed, the error
curves for GRUs with $d\geq2$ all coincide. This observation, together with the
limited performance gain from increasing $W$ in the AR($W$) and CNN($W$) models,
suggest that the error is approaching the Bayes limit. Note that the rate of
this convergence depends on $k$, since the size of the integration window
required to achieve the optimal performance depends on the degree of
non-Markovianity of the sequence {(see \cref{fig:error-vs-k}).} While we focused on comparing the performance of the models here, it is important to accurately compare the computational cost of these models as well. For a detailed comparison of the complexity we refer the reader to~\cref{app:complexity}. 

\subsection{The interplay between sequence memory and model architecture}

Finally, we found that for all three architectures -- AR, CNN and GRU -- there
is a peak in the reconstruction error, usually around $k\approx 5$. This peak
can be understood by noting that the tokens $x_t$ get more concentrated around
the origin as $k$ is increased, as we have shown in
\cref{eq:correlations}. Therefore, $x_t$ is less correlated with $c_t$ in the
highly non-Markovian limit. However, as the degree of non-Markovianity is
increased through increasing $k$, the history of the tokens position $x_t$ can
help with predicting the trap's position $c_t$ more accurately. The trade-off
between these two phenomena is evident in the non-monotonic dependency of the
error as a function of $k$ for models with larger integration window in
\cref{fig:error-vs-k}. For small $k$, the correlation between different times in
the sequence is not strong enough to compensate the loss of information about
the trap's position in the signal, but for larger $k$ the trap's position is
more regular and the history of $x_t$ can be used to infer the $c_t$ more
efficiently.

\section{Concluding perspectives}%
\label{sec:discussion}

We have applied the stochastically switching Ornstein-Uhlenbeck process
(SSOU) to model seq2seq machine learning tasks. This model gives us precise
control over the memory of the generated sequences, which we used to study the
interaction between this memory and various neural network architectures trained
on this data, ranging from simple auto-regressive models to convolutional and
recurrent neural networks. We found that the accuracy of these models is
governed by the interaction of the different time scales of the data, and we
discovered an intriguing, non-trivial interplay between the architecture of the students and the
sequence which leads to non-monotonic error curves.

An important limitation of the SSOU model considered here is that it only allows one to study correlations that decay exponentially fast. As a next step it would be interesting to explicitly consider the impact of long-range memory in the data sequence. 
This could be implemented in the SSOU model e.g.\ by choosing a long-ranged waiting time distribution, such as a power law. Another limitation concerns the linearity of the model analysed here. We speculate that a nonlinear model could provide insights into the different ways AR and CNN utilise the data. In our model, one could implement a tunable nonlinearity by considering e.g. a double-well potential with  increasing amplitude of the potential barrier. 
Similarly, to enrich the statistical dependencies within the sequence, one could generate non-symmetric sequences with respect to a $x \to -x $ sign inversion, by implementing two different waiting-time distributions. It would also be interesting to analyse the impact of sequence memory on the dynamics of simple RNN models~\cite{sompolinsky1988chaos, sussillo2009generating}, such as those with low-rank connectivity~\cite{mastrogiuseppe2018linking, schuessler2020interplay}.

Another exciting avenue would be to apply neural networks to noisy non-Markovian
signals extracted from experiments in physical or biological
systems~\cite{mindlinNonlinearDynamicsStudy2017, vettorettiFastPhysicsSlow2018,
  cavallaro2019effective,roldan2021quantifying, belousov2020volterra,
  Bruckner2019, skinner2021estimating}. Examples include the recent application
of the SSOU to infer the  
mitigation of the effects of non-Markovian
noise~\cite{mavadia2017prediction,majumder2020real} and the underlying heat dissipation of spontaneous oscillations of the
hair-cell bundles in the ear of the bullfrog~\cite{tucci2022modeling}. Applying our techniques to
such a biological system could be fruitful to decipher the hidden mechanisms and
statistics of switching in hearing and infer thermodynamic quantities beyond 
energy dissipation.



%% file: acknowledgements.tex
We thank Roman Belousov, Alessandro Ingrosso, Stéphane d'Ascoli, Andrea Gambassi, Florian Berger, Gogui Alonso, AJ Hudspeth, Aljaz Godec and Jyrki Piilo for stimulating discussions. A.S. is supported by a Chicago Prize Postdoctoral Fellowship in Theoretical Quantum Science. S.G. acknowledges funding from Next Generation EU, in the context of the National Recovery and Resilience Plan, Investment PE1 – Project FAIR “Future Artificial Intelligence Research”. This resource was co-financed by the Next Generation EU [DM 1555 del 11.10.22]. ER acknowledges financial support from PNRR MUR project PE0000023-NQSTI.

%% file: appendix.tex
\section{Analytical details}\label{sec:app-analytical}


\subsection{Details on the calculation of the phase diagram}
\label{sec:app-phase-diagram}

As anticipated in section \ref{sec:quantifying-non-markiovanity}, the process $C_t$ becomes Markovian in the case of exponentially distributed waiting time distribution; in our units this coincides with $\psi_{k=1}(\tau)=e^{-\tau}$.
For this simple case, one can characterise analytically the mono--bistable transition of the stationary density $p(x_t)$. This can be done by looking at the behaviour of $p(x_t)$ at the origin: if $x_t=0$ is a point of maximum, $p(x_t)$ is unimodal, bimodal otherwise.
We report from \cite{tucci2022modeling} the explicit expression of $p(x_t)$, which reads
\begin{equation}
    p(x_t)=\frac{1}{\sqrt{\pi}}\frac{\Gamma\left(\zeta+\frac{1}{2}\right)}{\Gamma\left(\zeta\right)}\int_{-1}^{+1}\mathrm{d}z \,\frac{e^{-\chi(x_t-z)^2}}{\sqrt{\pi/\chi}} (1-z^2)^{\zeta-1},
\end{equation}
where we have set $C_0=1$, and we have defined the dimensionless parameters $\zeta=t_\kappa/\langle\tau\rangle_k=1/\kappa$ and $\chi=t_{\rm diff}/t_\kappa=\kappa/(2D)$.
One finds that $x_t=0$ is an extremum point for $p(x_t)$, coinciding with the condition $p'(0)=0$. The nature of $x_t=0$ is understood by looking at the second derivative of $p(x_t)$, that is
\begin{equation}
p''(0)=\frac{2}{\sqrt{\pi/\chi^3}}\left[\frac{\chi}{\zeta+1/2}\,\,{}_1F_1\left(\frac{3}{2},\zeta+\frac{3}{2},-\chi\right)-{}_1F_1\left(\frac{1}{2},\zeta+\frac{1}{2},-\chi\right)\right],
\end{equation}
where ${}_1F_1$ denotes the confluent hypergeometric function.
The mono-bimodal transition occurs upon crossing the critical value $\chi^*$ satisfying the condition $p''(0)=0$, or equivalently, eq. \eqref{eq:chistar}.
For $\chi<\chi^*$, the second derivative $p''(0)$ is negative and $p(x)$ is unimodal, while it is bimodal otherwise.

When waiting times do not follow an exponential distribution, the system ceases to be Markovian. As far as our understanding goes, it becomes impossible to precisely compute the stationary probability density $p(x_t)$. Nevertheless, in the upcoming section, we will demonstrate that when $k>1$, it remains feasible to derive a complete expression for the two-point time correlators.

\subsection{Computation of the correlation functions in the non-Markovian case}
\label{sec:app-correlations}
\begin{figure}
  \centering
  \includegraphics[width=.6\columnwidth]{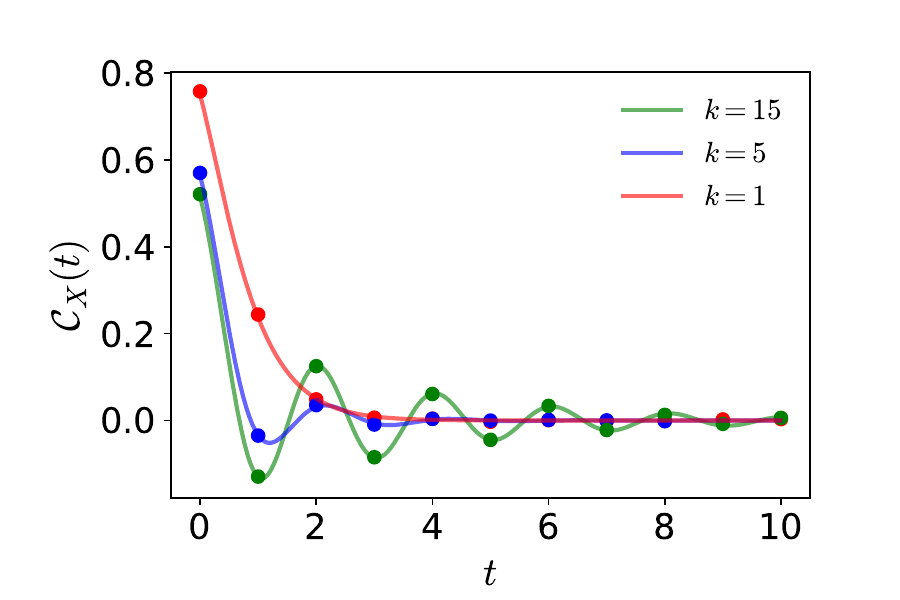}
  \caption{\label{fig:Cx}  \textbf{Autocorrelation function of the process $C_t$.} We compare the analytical formula of $\mathcal{C}_X(t)$ in eq. \eqref{eq:Cx} (solid line) with its numerical prediction (circles), calculated by generating $N=5\times10^4$ trajectories via Euler–Maruyama method with $\Delta t = 0.01$, $\kappa=2$, $D=0.5$ and various values of $k$.}
\end{figure}

In this section, we report the analytical expression for the (stationary) auto-correlation function $\mathcal{C}_X(t)$ of the process $X_t$ in eq. \eqref{eq:eom-x}, and its Fourier transform, i.e., the power spectral density (PSD) $S_X(\omega)$.
For Gamma-distributed waiting-time $\psi_k(\tau)$ as given in eq. \eqref{eq:waitingtimes}, the PSD $S_X(\omega)$ of the process $X_t$ is given by~\cite{tucci2022modeling}
\begin{equation}
S_X(\omega)=\frac{2D+\kappa^2S_C(\omega)}{\kappa^2+\omega^2},
\end{equation}
where $S_C(\omega)$ denotes the PSD of the process $C_t$, whose explicit expression reads
\begin{equation}
S_C(\omega)=\frac{4C_0^2}{\omega^2}\frac{R^2(\omega)-1}{R^2(\omega)+1+2R(\omega)\cos\phi(\omega)},
\end{equation}
with $R(\omega)=\left[1+(\omega/{k})^2\right]^{{k}/2}$, and $\phi(\omega)={k}\arctan(\omega/{k})$.
According to the Wiener-Khinchin theorem \cite{vankampen1992stochastic}, the inverse Fourier transform of $S_C(\omega)$ and $S_X(\omega)$ coincides with the auto-correlation functions $\mathcal{C}_C(t)\equiv \lim_{\tau\to\infty}\langle C_{t+\tau}\,C_\tau\rangle$ and $\mathcal{C}_X(t)\equiv \lim_{\tau\to\infty}\langle X_{t+\tau}\,X_\tau\rangle$. In the case of integer ${k}$, their expressions can be calculated explicitly according to
\begin{align}\label{eq:Cx}
&\mathcal{C}_C(t)=C_0^2\sum_{n=0}^{{k}-1} Q_n \,e^{-t\, {k}\, q_n},\\
&\mathcal{C}_X(t)=\frac{D}{\kappa}\, e^{-\kappa\,t}+C_0^2\,\kappa\sum_{n=0}^{{k}-1}Q_n\,\frac{{k}\, q_n\,e^{-\kappa\,t}-\kappa\,e^{-{k} \,q_n\,t}}{({k}\,q_n)^2-\kappa^2}, \label{eq:variance-steady-state}
\end{align}
where $Q_n\equiv -4(1-q_n)/({k}\,q_n)^2$, and $q_n\equiv 1-e^{i\pi(1+2n)/{k}}$ with $n=0,\dots,{k}-1$; in fig. \ref{fig:Cx}, we compare  eq. \eqref{eq:variance-steady-state} with the numerical estimate of $\mathcal{C}_X$ for three different choices of $k$. Note that the value of the sum $\sum_{n=0}^{{k}-1} Q_n =1$ implies the correct initial condition $C_c(0)=C_0^2$.
Moreover, one can deduce the stationary expression of the stationary variance $\langle x_t^2 \rangle$ of the process $X_t$ from the initial value $\mathcal{C}_X(0)$. 
We conclude this section by reporting the value of the covariance $\langle x_tc_t \rangle$, which we derive here using similar methods  as those in ref.~\cite{tucci2022modeling}, and is given by
\begin{equation}
  \langle x_t c_t \rangle =1-\frac{2}{\kappa}\frac{(1+\kappa/k)^{k}-1}{(1+\kappa/k)^{k}+1}.
\end{equation}

\subsection{Complexity of the models}\label{app:complexity}
In this section we analyse the size and computational complexity of the models used in this study. We note that the exact complexity depends on the specific implementation of the models. Therefore, we first analyse the scaling of the model complexity with respect to the number of model's memory units and then characterise the number of floating-point operations  (FLOPs) in our particular implementation. 

In this scaling analysis we do not consider the scaling with the sequence size it is the same for all of the models. The auto-regressive (AR) models we examine feature a 1D kernel of size $W$. Both the model's parameter count and computational complexity for its forward pass increase linearly with $W$. Similarly, 1D convolutional neural networks (CNN) comprise $f$ kernels of size $W$. For these models, the number of parameters and computational cost scale primarily as $Wf$.
The gated recurrent units (GRU) have a hidden state of size $d$. The number of parameters and the computational cost of these model scale with $d^2$. 

\begin{figure}
    \centering
    \includegraphics[width=0.6\columnwidth]{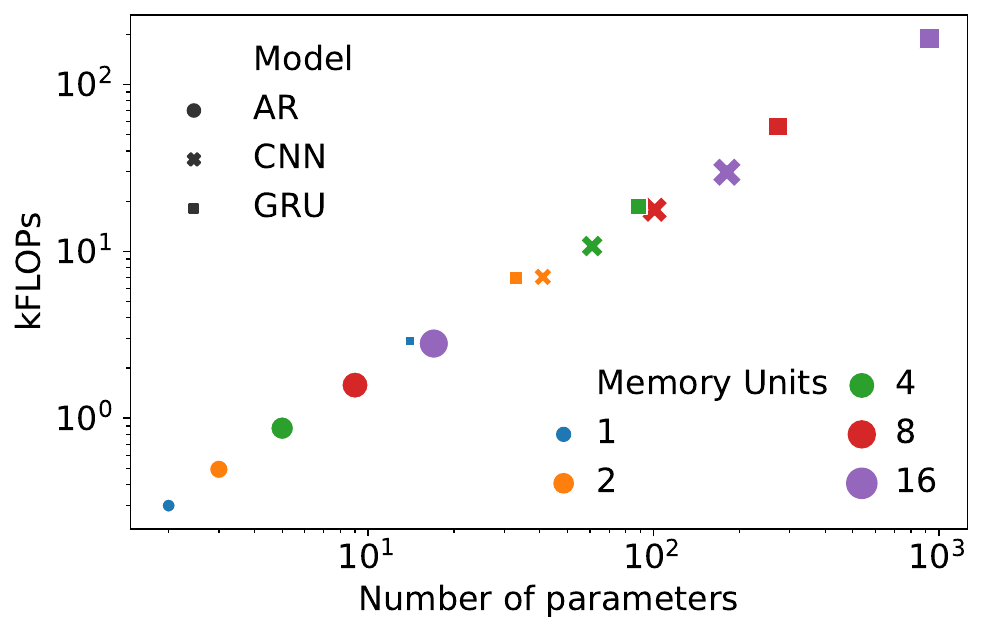}
    \caption{\textbf{Complexity of the models.} We analyze the size and computational complexity of the models used to generate~\cref{fig:error-vs-k}, i.e., auto-regressive (AR), convolutional (CNN), and gated recurrent neural network (GRU) architectures. Computational complexity is measured in terms of floating-point operations (FLOPs). The 'memory units' in these architectures correspond to the kernel size ($W$) in AR and the GRU hidden state size ($d$). Notably, all CNN models employ 10 filters in their first layer.}
    \label{fig:complexity}
\end{figure}

We employ software profiling to accurately compare the size and computational cost of the models used to produce~\cref{fig:error-vs-k}. As demonstrated in~\cref{fig:complexity}, our analysis agrees  with the numerical results. The AR models exhibit the smallest size and computational cost, followed by CNNs, and finally RNNs. Additionally, our simple analysis correctly predicts linear scaling of cost with the number of parameters.

\section{Additional experimental results}\label{sec:app-additional-experiments}

We use a fixed number of 40 epochs and do not perform any hyperparameter tuning (with the exception of the GRU(1) model as detailed in Appendix~\ref{sec:app-additional-experiments}) as the models are rather simple
and varying these parameters have negligible effects on the results.

\subsection{Additional details on creating the figures}

In this section, we provide additional details on the generation of some of the figures of the main text.

\paragraph{\Cref{fig:eq-phase-diagram}} The parameter sweep in the top left and top right panels
    was done by inspecting the intervals $\kappa \in [2/3,10]$ and
    $D\in [1/21,50] $ in a regular grid of size 20$\times$20.  We used a
    finite-sample estimate for Sarle's coefficient given by
    $(\sigma^2 + 1) / (3 + \omega \mathcal{K} ) $, with $\sigma$ the skewness
    and $\mathcal{K}$ the kurtosis of the sequence $\bX$, and
    $\omega = (T-1)^2/(T-2)(T-3)$ and $T=\tau/\Delta t = 3\times
    10^3$.

\subsection{Ensuring the consistency between continuous-time theory and sampled sequences}
\label{app:correlation-check}

Here we discuss how we ensured that our theory that is based on the continuous-time description of the stochastic processes $X_t$ and $C_t$ (cf.~\cref{sec:datamodel}) and the sequences we find into the machine learning models are consistent. If the sequence is sampled with a very small time step, 
the student almost never sees a jump of the trap, and effectively samples from an (equilibrium) distribution in one trap. On the other hand, in the  limit of very large time steps, the temporal correlations in the sequence is not visible, e.g., recall that the non-Markovianity parameter eventually decays to zero for infinitely large time differences. Thus, it is crucial to sample with a time step that is smaller but  comparable to the characteristic time scales of the data sequence. In practice, one would thus need to estimate those time scales in a preceding data analysis step, before feeding the data into the network. For this purpose, a suitable method is to compute the autocorrelation function of the input sequence data which reveals the time scales. For the SSOU model, we provide the analytical expression for the correlation function in \cref{eq:correlations}. For the SSOU, the characteristic time scales are given by the oscillation period and the decay time of the autocorrelation function, which are of order of magnitude 1  for our  parameter choice (see \cref{fig:Cx}
).

\subsubsection{Correlations function and integration time step}

To generate trajectories with correct statistics we vary the time differences $\Delta t$, which corresponds to the discrete version of $\mathrm{d} t$ in numerically integrating eq.~\eqref{eq:eom-x}. We find that $\Delta t= 0.005$ reproduces the correct statistics in fig.~\ref{fig:Cx}.

\subsubsection{The role of subsampling}
\label{app:subsampling}

As mentioned earlier the trajectories are first generated by numerically integrating eq.~\eqref{eq:eom-x} by choosing a sufficiently small $\mathrm{d} t$ such that the statistical properties of the trajectories match their theoretical values, which we verified by checking the correlation functions estimated analytically match our theoretical result
(see fig.~\ref{fig:Cx}).

As we discuss in the main text, we subsample the full trajectories before training the machine learning models. This subsampling also simplifies the optimization and accelerates the training. However, in doing so, we need to make sure that the qualitative properties of the model remain unchanged. Therefore, we compare AR models trained on the original trajectories with those trained on subsampled trajectories. We observe that the key properties of the problem, namely the strictly increasing errors with $k$ in the memoryless learning ($W=1$), and non-monotonic behaviour of errors due to  the trade-off between the memory ($W$) and the non-Markovianity ($k$) in more complicated models are preserved. 

\begin{figure}
    \centering
    \includegraphics[width=0.75\columnwidth]{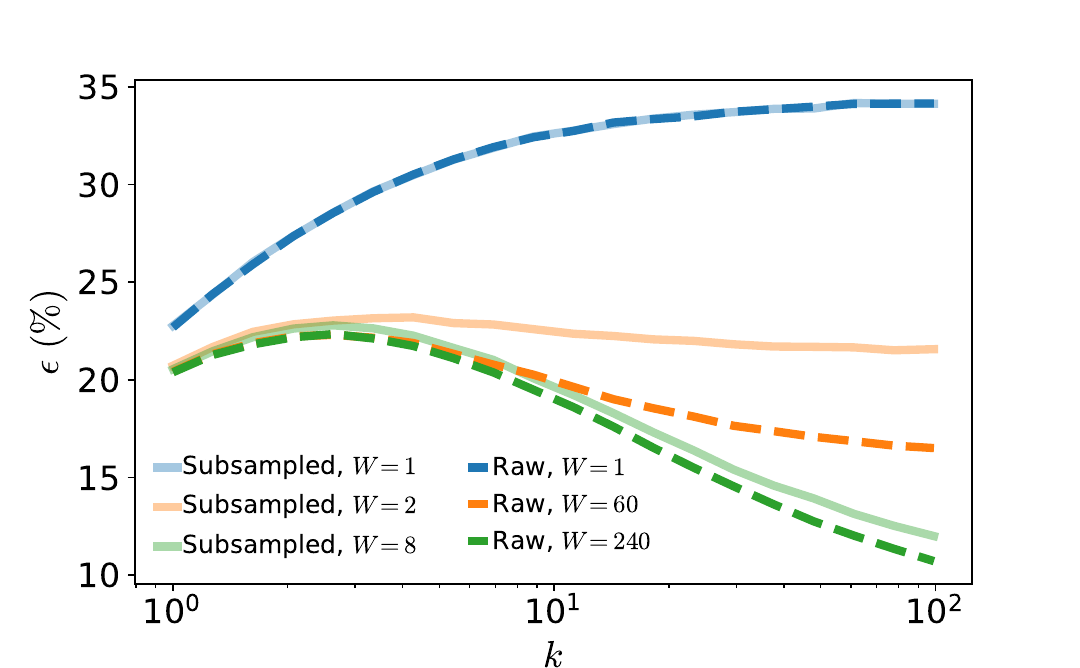}
    \caption{\textbf{Subsampling effects.} We compare the error of AR models trained on original raw dataset and the subsampled one with a subsampling factor of $s=30$. The key qualitative features of the error curves as function of the non-Markovianity $k$ are not affected by subsampling.}
    \label{fig:fine_vs_coarse}
\end{figure}

As mentioned in the main text we subsample the sequences by taking every $s$th element of the original sequence in the dataset when training the models in Sec.~\ref{sec:interplay_memory}. To investigate the effect of this subsampling we compare AR models trained on original dataset with those trained on subsampled sequences with $s=30$, and show the results in fig.~\ref{fig:fine_vs_coarse}. The qualitative features of the error curves as a function of non-Markovianity $k$ are preserved. The monotonicity of errors for memoryless models and the trade-off between improvements by memory and the difficulty of correlating the particle's position $x_t$ to the trap's position $c_t$, the two main features of fig.~\ref{fig:error-vs-k}, are present in both the original and the subsampled cases. 

\subsection{The thresholding algorithm}
\begin{figure}
    \centering
    \includegraphics[width=0.75\columnwidth]{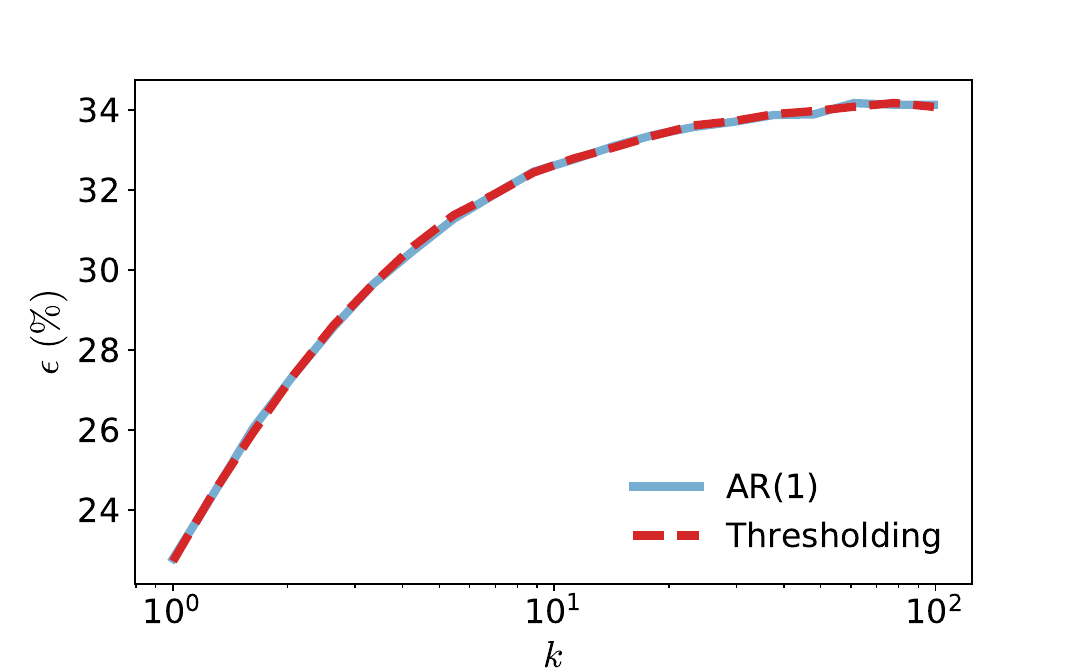}
    \caption{\textbf{Comparing the AR(1) and the thresholding algorithm.} Both the AR(1) model and the thresholding algorithm only have access to the position of the particle at a single point. As a consistency check we compare the performance of the two and observe that they match.}
    \label{fig:ar_vs_naive}
\end{figure}

As a baseline benchmark, we consider a simple thresholding algorithm, where $\hat c_t = \mathrm{sgn}(x_t)$. Put simply, this algorithm infers the potential's location by only considering the position of the particle at a single point $x_t$ and choose the closest configuration of potential $c_t$ to that position. In fig.~\ref{fig:ar_vs_naive} we compare the performance of the AR(1) model with the thresholding algorithm and observe that AR(1) matches the performance of the the thresholding algorithm. 
\subsection{Selecting the best GRU(1) model}
\begin{figure}
    \centering
    \includegraphics[width=0.75\textwidth]{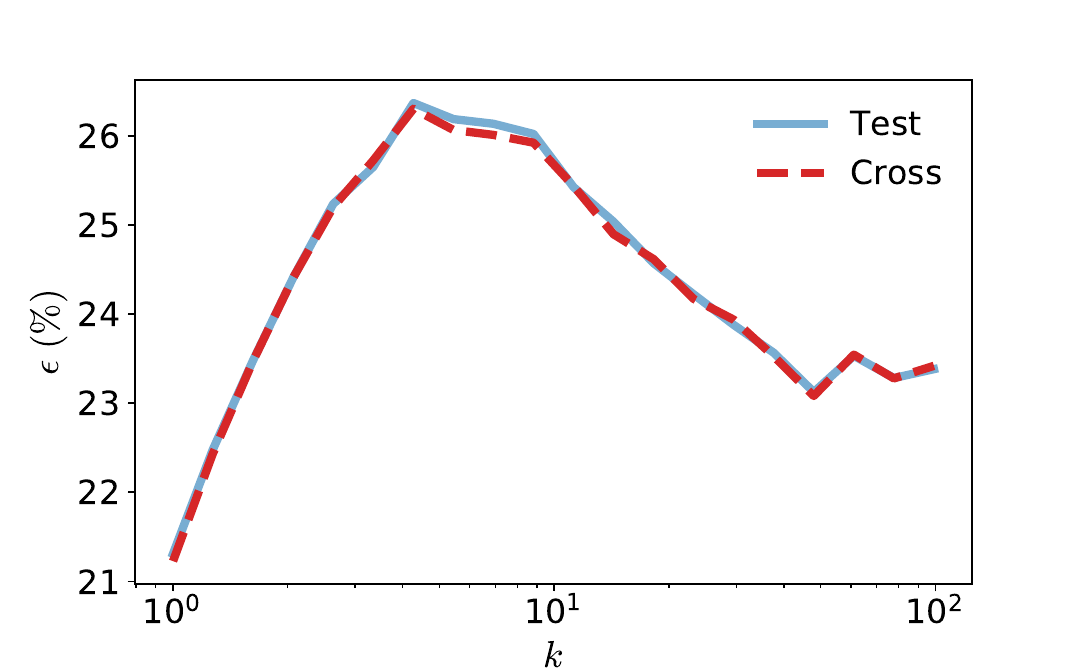}
    \caption{\textbf{Validating the GRU(1) model.} The accuracy of the best performing GRU(1) models over the cross validation and test sets match.} 
    \label{fig:rnn_cross}
\end{figure}

To find the best performing GRU(1) we train  model for all $k$ values shown in fig.~\ref{fig:error-vs-k} we train five different instance of the network initialized randomly using as suggested in Ref.~\cite{glorot2010understanding}. We choose the best performing model over 10000 samples for each $k$, and report the accuracy on a new set of 10000 samples. As shown in fig.~\ref{fig:rnn_cross} the test and cross validation accuracy are in excellent agreement. 
